\documentclass{article}
\usepackage{spconf,amsmath,graphicx,hyperref}
\usepackage{soul}
\usepackage{multirow}
\usepackage{xcolor,colortbl}
\definecolor{myorange}{HTML}{FF8000}
\usepackage{multirow}
\usepackage{booktabs}
\usepackage{graphicx} 
\usepackage[ruled,vlined]{algorithm2e} 
\usepackage{enumitem}
\usepackage[T1]{fontenc}

\title{RAVE: Retrieval and Scoring Aware Verifiable Claim Detection}
\name{Yufeng Li, Arkaitz Zubiaga}
\address{Queen Mary University of London, London, UK}
\begin{document}
%
\maketitle
\begin{abstract}
The rapid spread of misinformation on social media underscores the need for scalable fact-checking tools. A key step is claim detection, which identifies statements that can be objectively verified. Prior approaches often rely on linguistic cues or claim check-worthiness, but these struggle with vague political discourse and diverse formats such as tweets. We present RAVE (Retrieval and Scoring Aware Verifiable Claim Detection), a framework that combines evidence retrieval with structured signals of relevance and source credibility. Experiments on CT22-test and PoliClaim-test show that RAVE consistently outperforms text-only and retrieval-based baselines in both accuracy and F1.
\end{abstract}
\section{Introduction}
\label{sec:intro}

On social media, misinformation spreads more quickly and widely than in traditional formats, and the content is highly diverse, ranging from news and political debates to online posts such as tweets \cite{zubiaga2018detection, oladokun2024misinformation}. Such content can shape public opinion and cause societal harm, making fact-checking increasingly important. Traditional fact-checking, however, is labor-intensive and time-consuming, which has led to growing interest in Automated Fact-Checking Systems (AFC) that leverage artificial intelligence to support and scale this process \cite{johnson2024case}. Given the scale and speed of misinformation, claim detection serves as a crucial first step. It identifies factual statements that require verification, filtering verifiable claims for the downstream task of claim verification, where they are ultimately fact-checked as true or false.

The task of claim detection has been defined in different ways, with most studies focusing on claim check-worthiness \cite{kartal2023ReThinkYou}, where claims are prioritized based on criteria such as public importance or interest \cite{panchendrarajan2024claim}. More recent work shifts toward verifiable claim detection, which defines a claim as an assertion that can be objectively checked \cite{konstantinovskiy2021AutomatedFactchecking, ni2024afacta}. In line with this view and inspired by the downstream task of claim verification, we define \textbf{a verifiable claim as one that contains at least one factual statement that can be objectively verified through external evidence}.

Early approaches to claim detection relied on traditional machine learning systems such as ClaimBuster \cite{hassan2017ClaimBusterfirstever}, ClaimRank \cite{jaradat2018ClaimRankDetecting}, and the CNC system \cite{konstantinovskiy2021AutomatedFactchecking}. With the rise of deep learning, the CheckThat! shared tasks introduced LSTM-based models \cite{dhar2019hybrid}, but transformer-based architectures have dominated since 2020 \cite{williams2020accenture}. More recently, large language models (LLMs) have driven further progress: top systems in CheckThat! 2023 used GPT-3 in zero-shot and few-shot settings \cite{sawinski2023openfact}, and the 2024 winner fine-tuned eight open-source LLMs \cite{li2024factfinders}. AFaCTA extended this line of work with a multi-step prompting framework that improved factual claim annotation \cite{ni2024afacta}.

These methods have been evaluated on two main datasets: sentence-level political debate transcripts and COVID-19 tweets independently. Debate transcripts are often short, vague, and increasing ambiguity, while tweets are more self-contained and explicit. For example, debates may include statements like ``the last couple of years have been especially trying for our medical professionals", which lack clear reference points, whereas tweets such as ``Pfizer isn’t Lamborghini. Sinovac isn’t Proton" are more concrete. Most research has concentrated on debates \cite{konstantinovskiy2021AutomatedFactchecking, alam2023overview, ni2024afacta}, with fewer studies addressing tweets \cite{shaar2021overview, nakov2022overview}.

To address these limitations, we propose RAVE (Retrieval-and-Scoring-Aware Verifiable Claim Detection), a framework that improves verifiable claim detection by incorporating evidence retrieval and scoring into the decision process. Our intuition is that the verifiability depends not only on its linguistic form but also on the availability and quality of external evidence that can be used in the downstream task. We evaluate our approach on the CT22-test~\cite{nakov2022overview} and PoliClaim-test~\cite{ni2024afacta} datasets, comparing it with standard text-only baselines and retrieval-based variants to isolate the role of scoring and selection strategies. Our main contributions are:

\begin{itemize}[leftmargin=*,noitemsep,topsep=0pt]
\item We provide a precise formulation of claim detection as the task of identifying \textbf{verifiable claims}, defined as factual statements that can be objectively checked against external evidence, regardless of their truth value.
\item We introduce \textbf{RAVE}, a retrieval- and LLM-based framework that integrates evidence retrieval with relevance and credibility scoring, using these scores for both evidence selection and verifiability estimation. Code, scripts, logs, and processed datasets will be publicly released upon acceptance.
\item We conduct a systematic comparison of claim detection across datasets with different linguistic and structural properties, providing empirical insights into how dataset characteristics affect verifiability modeling.
\end{itemize}


\section{Proposed Method}
\label{sec:method}

\begin{algorithm}[t]
\caption{\textbf{RAVE}: \textbf{R}etrieval-and-Scoring-\textbf{A}ware \textbf{V}erifiable Claim D\textbf{E}tection}
\label{alg:rave}
\KwIn{Claim $x$}
\KwOut{Verifiability label $y \in \{\text{VERIFIABLE}, \text{NON-VERIFIABLE}\}$}

\BlankLine
\textbf{Step 1: Entity-based retrieval} \\
Extract entities $E = \{e_1, \dots, e_m\}$ from $x$ (PERSON, ORG, LOCATION, EVENT, CLAIM\_OBJECT). \\
For each $e_i \in E$, query a web search API and collect candidate snippets 
$S = \{s_1, \dots, s_n\}$ with metadata (domain, title, url). 

\BlankLine
\textbf{Step 2: Scoring} \\
For each snippet $s_j \in S$: \\
\Indp
    Compute relevance $r_j = \cos(\mathbf{h}_x, \mathbf{h}_{s_j})$ between embeddings. \\
    Compute credibility $c_j$ from source domain metadata. \\
    
\Indm

\BlankLine
\textbf{Step 3: Scoring-aware selection} \\
Aggregate score: $score_j = \alpha r_j + (1-\alpha) c_j$. \\
Select top-$K$ snippets $S_K \subseteq S$ according to $score_j$. 

\BlankLine
\textbf{Step 4: Verifiability decision} \\
Build a structured prompt including: \\
\Indp
- Claim $x$ \\
- Selected snippet contents, metadata (domain, title) \\
- Relevance and credibility scores of each snippet. \\
\Indm

Feed the prompt to an LLM. \\
Obtain output $y \in \{\text{VERIFIABLE}, \text{NON-VERIFIABLE}\}$.

\end{algorithm}

To move beyond simple evidence retrieval, we propose a scoring-aware evidence selection and verifiability decision framework for claim detection, called RAVE (Retrieval-and-Scoring-Aware Verifiable Claim Detection), as shown in Algorithm \ref{alg:rave}.

\subsection{Entity-based Retrieval}

To retrieve initially relevant evidence, we first perform entity extraction on the input claim. Entities are short and informative units that make a sentence more suitable for linking to external knowledge. We define five types of entities: PERSON, ORG, LOCATION, EVENT, and CLAIM\_OBJECT. The latter two types capture specific occurrences and central objects or concepts in a claim, which extend beyond traditional entity definitions. Entities are extracted using a zero-shot learning approach with prompt-based instructions.

Once entities are extracted, we query the Google API with each entity to retrieve candidate snippets, denoted as $S = \{s\_1, \dots, s\_n\}$, along with their metadata. These snippets are used to construct a context pool for the claim. We retain snippets rather than full web pages, since the goal is not to gather comprehensive evidence for verification but to provide contextual signals that help determine whether a claim is verifiable.

\subsection{Scoring}

\begin{table}[t]
\centering
\begin{tabular}{cp{0.7\columnwidth}}
\toprule
Score & Domain type / Examples \\
\midrule
1.00 & Highly authoritative (Wikipedia, Reuters, BBC, Nature) \\
0.95 & Government (.gov), educational (.edu) \\
0.85 & Academic and research institutions (universities, institutes) \\
0.75 & Established news outlets (news, times, post, journal) \\
0.65 & Non-profit organizations (.org) \\
0.50 & General commercial (.com) \\
0.40 & Other or unclassified domains \\
\bottomrule
\end{tabular}
\caption{Heuristic credibility scores by domain type.}
\label{tab:credibility-scores}
\end{table}

To evaluate the usefulness of retrieved snippets, we assign two scores. Relevance is measured by cosine similarity between a snippet and the input claim, while source credibility reflects the reliability of the snippet’s domain following a source-centric view of information quality \cite{pasi2020information}. Each domain is assigned a normalized score between 0.4 and 1.0 (Table~\ref{tab:credibility-scores}), which estimates the general reliability of a source rather than the truthfulness of individual claims, consistent with prior fact-checking and information retrieval research \cite{westerman2014social}.

\subsection{Scoring-aware Selection}

We then compute a combined score for each snippet: 
\vspace{-0.5em}
\[
score_j = \alpha r_j + (1-\alpha) c_j
\vspace{-0.5em}
\]

\noindent The parameter $\alpha$ is tuned on the CT22-dev dataset through a coarse grid search over $\{0.3,\ldots,0.8\}$, and $\alpha=0.6$ is selected using the one-standard-error rule; no additional tuning is applied to other datasets or models. This score balances snippet relevance ($r\_j$) with source credibility ($c\_j$), supporting snippets that are both contextually aligned and reliable. In this way, the retrieved context reduces noise and misinformation while supporting effective claim detection.

\subsection{Verifiability decision}

In the final decision step, RAVE treats retrieval signals as structured indicators of verifiability rather than as direct evidence for truth assessment. For each selected snippet, the LLM receives i) its textual content, ii) the computed relevance score, iii) the source credibility score, and iv) metadata such as the domain. Providing these signals explicitly allows the model to evaluate not only the retrieved information but also its reliability and alignment with the claim. Hence, instead of assessing whether the claim is formulated in a way that can be objectively checked given the availability of relevant and credible external information, the decision thus depends on the quality and reliability of potential evidence, aligning RAVE with the goal of detecting verifiable claims.

\section{Experiments}
\label{sec:experiment}

\subsection{Datasets and Baselines}\label{subsec:datasets_baselines}

We evaluate our framework on two benchmark datasets: CT22-test and PoliClaim-test. CT22-test consists of COVID-19 tweets that are typically longer, self-contained, and information-rich, with explicit references to entities, events, and time. In contrast, PoliClaim contains debate transcripts segmented into single-sentence units that are often short, ambiguous, and context-dependent, frequently relying on pronouns and rhetorical devices.

All methods use GPT-4o as the backbone LLM with the same prompt template; only the retrieval strategy varies. The baselines are designed to isolate the contribution of retrieval and scoring to verifiability detection.

\begin{itemize}[leftmargin=*,noitemsep,topsep=0pt]
    \item Text-only: The model determines the verifiability using only the input claim, without retrieved context. This serves as the standard baseline in prior claim detection work.
    \item Rand-K: A random set of $k$ snippets is added as context, providing a lower bound that controls for the effect of extra text without using relevance.
    \item Search-K: The top-$k$ snippets are selected according to the search engine’s default ranking, testing the impact of naive search-based retrieval.
    \item RAVE-Stats: The model receives only aggregated retrieval statistics, such as entity coverage, snippet coverage, source diversity, and inter-snippet agreement, to assess whether statistical indicators alone can determine verifiability.
    \item RAVE-Meta: The model is given only snippet metadata, including domain, and associated relevance and credibility scores, without snippet text. This evaluates whether metadata and scoring signals are sufficient.
    \item RAVE (ours): Our full method, which combines snippet content with relevance and credibility scores, enabling the model to use both textual evidence and structured signals.
\end{itemize}

\subsection{Experiment Setup}

For reproducibility, all experiments use GPT-4o-08-06 with temperature set to 0, top-p to 1, and a maximum token length of 500. All baselines and our proposed method follow the same prompt template, differing only in input content and decision strategy, while outputs are consistently formatted in JSON. Performance is evaluated primarily using accuracy and F1-score, with precision and recall reported as supporting metrics. Results are presented for both CT22-test and PoliClaim-test.

\section{Results and Discussion}

\subsection{Main Results}

\begin{table*}[t]
\centering
\begin{tabular}{lcccc|cccc}
\toprule
\multirow{2}{*}{\textbf{Method}} & \multicolumn{4}{c|}{\textbf{CT22-test}} & \multicolumn{4}{c}{\textbf{PoliClaim-test}} \\
 & Acc. & Prec. & Rec. & F1 & Acc. & Prec. & Rec. & F1  \\
\midrule
Text-only   & 0.8207 & \textbf{0.8824} & 0.8054 & 0.8421 & 0.6789 & \textbf{1.0000} & 0.4971 & 0.6641 \\
Rand-K      & 0.8048 & 0.8333 & 0.8389 & 0.8361 & 0.6998 & \textbf{1.0000} & 0.5298 & 0.6926 \\
Search-K    & 0.8287 & 0.8487 &\textbf{0.8658} & \textbf{0.8571} & 0.6985 & \textbf{1.0000} & 0.5278 & 0.6910\\
RAVE-Stats  & 0.8088 & 0.8389 & 0.8389 & 0.8389 & 0.6446 & \textbf{1.0000} & 0.4434 & 0.6144\\
RAVE-Meta   & 0.8088 & 0.8435 & 0.8322 & 0.8378 & \textbf{0.7010} & \textbf{1.0000} & 0.5317 & 0.6942 \\
\textbf{RAVE} & \textbf{0.8327} & 0.869 & 0.8456 & \textbf{0.8571} &
\textbf{0.7010} & 0.9964 & \textbf{0.5336} & \textbf{0.6950} \\
\bottomrule
\end{tabular}
\caption{Results on verifiable claim detection. All methods use GPT-4o as the backbone LLM with fixed K=3; only the retrieval strategy differs. Best scores are highlighted in \textbf{bold}.}
\label{tab:main_results}
\end{table*}

As mentioned in Section \ref{subsec:datasets_baselines} and shown in Table~\ref{tab:main_results}, performance is generally higher on CT22-test, where retrieval provides sufficient context and all methods achieve balanced precision and recall. RAVE attains the best overall F1 by combining high precision with strong recall. PoliClaim-test is more challenging because of its higher entity sparsity. While only 4.7\% of verifiable and 24.5\% of non-verifiable claims in CT22 lack entities, the proportions in PoliClaim rise to 30.3\% and 81.0\%. This imbalance leads LLMs to default to the NON-VERIFIABLE label, producing very high precision but low recall for baselines. By incorporating source credibility, RAVE reduces this bias, improves recall while preserving precision, and achieves the highest overall F1. These results demonstrate RAVE’s robustness across datasets with different entity densities and contextual richness.

\subsection{Ablation Study}

\subsubsection{Effectiveness of Scoring Components}

\begin{table}[t]
\centering
\begin{tabular}{lccccc}
\toprule
Variant & Acc & Prec & Rec & F1 & $\Delta$F1 \\
\midrule
Relevance only & 80.9 & 83.9 & 83.9 & 83.9 & -1.8\\
Credibility only & 82.5 & 84.8 & \textbf{85.9} & 85.3 & -0.4 \\
Text+Snippets & 82.1 & 85.1 & 84.6 & 84.9 & -0.8 \\
\textbf{RAVE} & \textbf{83.3} & \textbf{86.9} & 84.6 & \textbf{85.7} & \textbf{+0.0} \\
\bottomrule
\end{tabular}
\caption{Ablation study of scoring components on CT22-test under fixed $K{=}3$ with identical prompts. We compare models using only \textit{relevance}, only \textit{credibility}, no scores, and their combination (\textbf{RAVE}). Best results are highlighted in \textbf{bold}.}
\label{tab:ablation-scores}
\end{table}

We conduct an ablation study on CT22-test with $K{=}3$, comparing relevance, credibility, raw text, and their combination in RAVE (Table~\ref{tab:ablation-scores}). Relevance alone yields high precision (83.9\%) but lower F1, as retrieved evidence may not always be reliable. Credibility achieves the highest recall (85.9\%) but lower precision, since trustworthy sources are not always directly relevant. Using only input text and snippets remains competitive (F1 84.9\%), but explicit scoring improves stability. RAVE, which combines relevance and credibility, achieves the best overall results (accuracy 83.3\%, precision 86.9\%, F1 85.7\%), showing that the two signals are complementary and most effective when used together.

\subsubsection{K-Sensitivity}

\begin{figure}[t]
    \centering
    \includegraphics[width=0.8\linewidth]{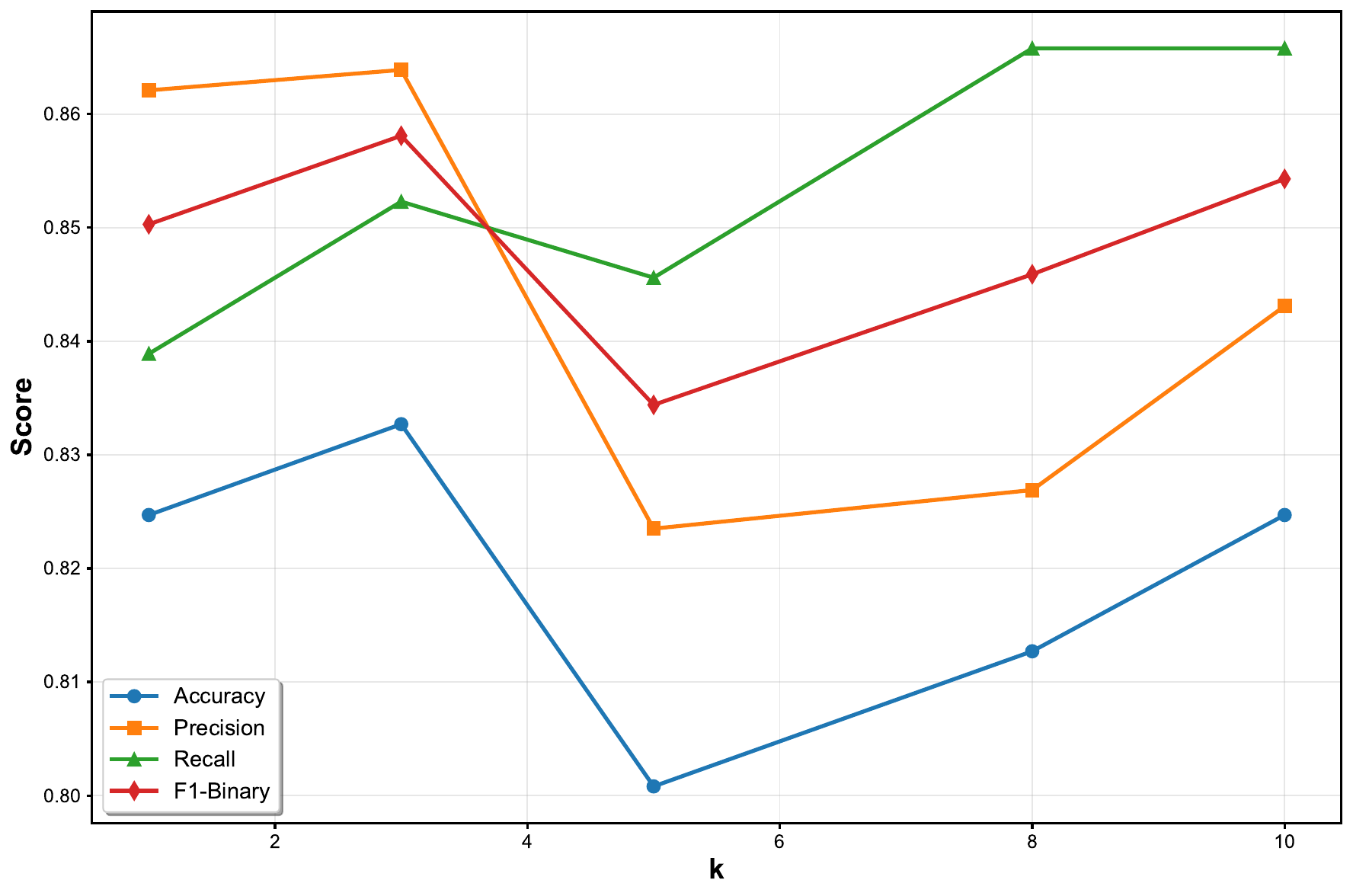}
    \caption{Sensitivity of model performance to the number of retrieved snippets ($K$) on CT22-test.}
    \label{fig:k-sensititvity}
\end{figure}

We vary the number of retrieved snippets $K \in \{1,3,5,8,10\}$ for RAVE (GPT-4o) on CT22-test. As shown in Fig.~\ref{fig:k-sensititvity}, recall increases as $K$ grows, while precision declines slightly, producing a shallow U-shaped F1 curve. The best performance occurs at $K=3$ (F1=85.7\%), with comparable results at $K=10$ (85.4\%). The dip at $K=5$ indicates added redundancy and noise, though the decline is modest due to the scoring signals in the prompt. For efficiency, we set $K=3$ for all subsequent experiments.

\subsection{Error Analysis}

To further understand the performance of RAVE, we conducted a qualitative error analysis using representative examples from the CT22 and PoliClaim datasets. The results reveal distinct error patterns shaped by the linguistic features of each dataset and the challenges of verifiable claim detection.

On CT22, the system achieved balanced precision and recall, yielding higher overall accuracy (83.3\%). Errors were more evenly distributed:

\begin{itemize}[leftmargin=*,noitemsep,topsep=0pt]
    \item False positives often involved non-verifiable claims expressed with factual phrasing. For example, ``As president, I will: - Implement nationwide mask mandates - Ensure access to regular, reliable, and free testing - Accelerate the development of treatments and vaccines I won’t waste any time getting this virus under control." lists specific actions but describes intent rather than verifiable facts.
    \item False negatives frequently arose from sarcasm or mixed fact–opinion constructions. For instance, ``KFC had 11 secret herbs and spices and nobody questioned it for years but suddenly you want to know what’s in the Covid vaccine." contains a verifiable historical reference but was misclassified due to its sarcastic tone.
\end{itemize}

On PoliClaim, the system struggled more, reflected in lower recall for verifiable claims. Errors showed systematic difficulties with political discourse:

\begin{itemize}[leftmargin=*,noitemsep,topsep=0pt]
    \item The only false positive came from rhetorical speech, e.g., ``After everything we’ve faced over the past three years, it’s my honor to report that not only is the State of our State Resilient, we’re fulfilling our motto of ‘North to the Future.’" Although symbolic, the model misinterpreted it as fact-based.
    \item Most false negatives involved broad generalizations lacking specificity. Statements like ``When our prison system went unaddressed for decades and resulted in serious challenges, we found a way toward a solution." and ``When our roads and bridges were in need for desperate improvements, we found a way to make significant progress all across the state." describe verifiable conditions but were treated as rhetorical because they lacked concrete details such as dates or entities.
\end{itemize}

\section{Conclusion}
\label{sec:typestyle}

In this work, we propose RAVE, a retrieval and scoring aware framework for verifiable claim detection that integrates external evidence with relevance and credibility signals. Experiments on CT22 and PoliClaim show consistent improvements over text-only and retrieval-based baselines in both accuracy and F1. While performance on PoliClaim is about 10\% lower due to ambiguity in debate transcripts, future work will focus on domain adaptation and refined credibility modeling to enhance robustness across diverse datasets.

\bibliographystyle{IEEEtran}
\bibliography{main}

\end{document}